\documentclass[a4paper,11pt]{article}
\usepackage{amsmath, amssymb, hyperref, graphicx, geometry, booktabs, array, caption, setspace}
\usepackage{lipsum}
\usepackage{natbib}
\title{Hybrid Group Relative Policy Optimization: A Multi-Sample Approach to Enhancing Policy Optimization}
\author{Soham Sane \\ \small Collins Aerospace}
\date{January 30, 2025}

\begin{document}

\maketitle

\begin{abstract}
Hybrid Group Relative Policy Optimization (Hybrid GRPO) is a reinforcement learning framework that extends Proximal Policy Optimization (PPO) and Group Relative Policy Optimization (GRPO) by incorporating empirical multi-sample action evaluation while preserving the stability of value function-based learning. Unlike DeepSeek’s GRPO, which eliminates the value function in favor of purely empirical reward estimation, Hybrid GRPO introduces a structured advantage computation method that balances empirical action sampling with bootstrapped value estimation. This approach enhances sample efficiency, improves learning stability, and mitigates variance amplification observed in purely empirical methods. A detailed mathematical comparison between PPO, DeepSeek GRPO, and Hybrid GRPO is presented, highlighting key differences in advantage estimation and policy updates. Experimental validation in a controlled reinforcement learning environment demonstrates that Hybrid GRPO achieves superior convergence speed, more stable policy updates, and improved sample efficiency compared to existing methods. Several extensions to Hybrid GRPO are explored, including entropy-regularized sampling, hierarchical multi-step sub-sampling, adaptive reward normalization, and value-based action selection. Beyond reinforcement learning in simulated environments, Hybrid GRPO provides a scalable framework for bridging the gap between large language models (LLMs) and real-world agent-based decision-making. By integrating structured empirical sampling with reinforcement learning stability mechanisms, Hybrid GRPO has potential applications in autonomous robotics, financial modeling, and AI-driven control systems, such as Tesla’s Full Self-Driving (FSD) and autonomous drone navigation. These findings suggest that Hybrid GRPO serves as a robust and adaptable reinforcement learning methodology, paving the way for further advancements in policy optimization.
\end{abstract}

\newpage

\tableofcontents

\newpage

\newpage

\section{Introduction}
\label{sec:intro}

Reinforcement learning (RL) has achieved significant success across various domains, from robotic control to game playing. Among policy optimization methods, Proximal Policy Optimization (PPO) \cite{schulman2017proximal} remains one of the most widely adopted techniques due to its balance between sample efficiency and stability. PPO leverages a value function \( V(s) \) to estimate expected returns, allowing for variance reduction in policy gradient updates. However, the reliance on bootstrapped value function approximation introduces bias, which can limit learning performance in certain settings.

\vspace{10pt}

Group Relative Policy Optimization (GRPO) was recently introduced by DeepSeek \cite{deepseek2025r1} as an alternative to value function-based approaches. GRPO eliminates \( V(s) \) entirely and instead computes empirical returns by sampling multiple actions per state. While this approach removes function approximation bias, it significantly increases sample complexity and introduces high reward variance, which may impede convergence in practical applications.

\vspace{10pt}

This work presents Hybrid Group Relative Policy Optimization (Hybrid GRPO), a reinforcement learning framework that combines the strengths of both PPO and GRPO. Hybrid GRPO retains the value function \( V(s) \) while integrating multiple action samples per macro-step, extracting additional training data without sacrificing the stability provided by value function-based bootstrapping. By incorporating systematic empirical sampling and adaptive reward transformations, Hybrid GRPO increases data efficiency while reducing the variance typically associated with purely empirical return-based methods.

\vspace{10pt}

\subsection{Contributions of This Work}

This paper introduces a new reinforcement learning paradigm and provides both theoretical and empirical validation of Hybrid GRPO. The key contributions of this work are as follows:

\vspace{10pt}

\begin{itemize}
    \item A novel mathematical formulation for advantage estimation that integrates empirical return sampling while preserving \( V(s) \), bridging the gap between PPO and DeepSeek GRPO.
    \item A multi-sample estimation strategy that improves data density by extracting multiple training samples per macro-step.
    \item A comprehensive mathematical comparison of PPO, DeepSeek GRPO, and Hybrid GRPO, detailing the differences in their advantage estimation methods, policy updates, and sample efficiency.
    \item Empirical validation through a custom synthetic simulation, demonstrating improved sample efficiency, faster convergence, and more stable policy updates compared to PPO and DeepSeek GRPO.
    \item A discussion on potential future enhancements, including entropy-enhanced sampling, multi-step sub-sampling, adaptive reward scaling, and learned value-based action selection to further refine the Hybrid GRPO methodology.
\end{itemize}

\vspace{10pt}

Hybrid GRPO represents a new direction in reinforcement learning by combining empirical action sampling with structured value function estimation, ensuring a balance between sample efficiency, training stability, and variance reduction. The findings of this work suggest that multi-sample advantage estimation provides a more robust learning signal compared to both PPO and DeepSeek GRPO, paving the way for improved reinforcement learning architectures.

\newpage

\newpage

\section{Mathematical Comparison: PPO vs. GRPO vs. Hybrid GRPO}

This section presents a detailed mathematical comparison of three reinforcement learning approaches. 

\begin{enumerate}
    \item Proximal Policy Optimization (PPO), which relies on a value function \( V(s) \) to estimate expected returns.
    \item DeepSeek’s Group Relative Policy Optimization (GRPO), which completely eliminates \( V(s) \) and replaces it with empirical return estimates from multiple action samples.
    \item The proposed Hybrid GRPO, which retains \( V(s) \) while incorporating multiple sampled actions per macro-step to enhance the policy update.
\end{enumerate}

Each approach is analyzed in terms of its formulation of the advantage function, action sampling strategy, and reward estimation methodology.

\subsection{Proximal Policy Optimization (PPO)}

PPO relies on a learned value function \( V(s) \) as a baseline for estimating the advantages. The advantage function is defined as:

\begin{equation}
    A_T = Q(s_T, a_T) - V(s_T)
\end{equation}

Expanding Q(s\_T, a\_T) using the Bellman equation \citep{sutton2018reinforcement}:

\begin{equation}
    A_T = \left[ R_T + \gamma V(s_{T+1}) \right] - V(s_T)
\end{equation}

where:
\begin{itemize}
    \item \( R_T = r(s_T, a_T) \) is the reward obtained from executing action \( a_T \) in state \( s_T \).
    \item \( V(s_T) \) is the estimated value function, computed using a critic network.
    \item \( V(s_{T+1}) \) is the predicted value of the next state.
    \item \( \gamma \) is the discount factor.
\end{itemize}

The PPO loss function optimizes a clipped surrogate objective:

\begin{equation}
    \mathcal{L}_{\text{PPO}} = \mathbb{E} \left[ \min \left( \rho_T A_T, \text{clip}(\rho_T, 1-\epsilon, 1+\epsilon) A_T \right) \right]
\end{equation}

where \( \rho_T \) is the probability ratio:

\begin{equation}
    \rho_T = \frac{\pi_{\theta}(a_T | s_T)}{\pi_{\theta_{\text{old}}}(a_T | s_T)}
\end{equation}

\subsection{DeepSeek GRPO (Empirical Return-Based)}

DeepSeek GRPO removes the value function entirely and instead estimates advantages using empirical action sampling \citep{deepseek2025r1}. At each macro-step \( T \), multiple actions are sampled:

\begin{equation}
    a_T^{(t)} \sim \pi_{\theta}(a | s_T), \quad t = 1, ..., N
\end{equation}

For each sampled action, an empirical reward is computed (Raw reward is normalized via a tanh() activation function):

\begin{equation}
    R_T^{(t)} = r(s_T, a_T^{(t)})
\end{equation}

The advantage function is then approximated as:

\begin{equation}
    A_T = \frac{1}{N} \sum_{t=1}^{N} R_T^{(t)} - \mathbb{E}[R_T^{(t)}]
\end{equation}

where the expectation term is the empirical mean of all sampled rewards:

\begin{equation}
    \mathbb{E}[R_T^{(t)}] = \frac{1}{N} \sum_{t=1}^{N} R_T^{(t)}
\end{equation}

The loss function for GRPO is given by:

\begin{equation}
    \mathcal{L}_{\text{GRPO}} = \mathbb{E} \left[ \min \left( \rho_T A_T, \text{clip}(\rho_T, 1-\epsilon, 1+\epsilon) A_T \right) \right]
\end{equation}

where the probability ratio \( \rho_T \) remains identical to PPO. The key difference is that GRPO does not rely on a critic function and instead derives all value estimates directly from sampled actions. 

\subsection{Hybrid GRPO: Multi-Sample Reward Estimation with \texorpdfstring{$V(s)$}{V(s)}}

Hybrid GRPO extends the PPO framework by preserving the value function while incorporating multi-sampling at each macro-step. The primary objective is to extract more informative data points per state transition while maintaining the stability benefits of bootstrapped value function estimation. The key difference with Hybrid GRPO is that raw rewards are compared locally at each macro time step, T, and therefore pass through an activation function tanh() prior to being tabulated as extra data points.

\text{}

\noindent \textbf{Step 1: Multi-Sampling Actions} \\
Instead of selecting a single action \( a_T \), Hybrid GRPO samples multiple actions per macro-step:

\begin{equation}
    a_T^{(t)} \sim \pi_{\theta}(a | s_T), \quad t = 1, ..., N
\end{equation}
\text{}

\noindent \textbf{Step 2: Computing Empirical Rewards} \\
For each sampled action \( a_T^{(t)} \), the environment's reward function is evaluated:

\begin{equation}
    R_T^{(t)} = r(s_T, a_T^{(t)})
\end{equation}
\text{}

\noindent \textbf{Step 3: Applying a Reward Transformation Function} \\
A function \( f(R) \) is introduced to modify sampled rewards, improving stability:

\begin{equation}
    \tilde{R}_T^{(t)} = f(R_T^{(t)})
\end{equation}

where \( f(R) \) is an adaptive transformation function that can normalize, clip, or scale rewards. The default normalization function is tanh().

\vspace{10pt} 

\noindent \textbf{Step 4: Computing Multi-Sample Advantage} \\
The advantage function incorporates multiple empirical samples while still leveraging \( V(s) \):

\begin{equation}
    A_T = \frac{1}{N} \sum_{t=1}^{N} \left[ \tilde{R}_T^{(t)} + \gamma V(s_{T+1}^{(t)}) - V(s_T) \right]
\end{equation}

This formulation introduces several improvements over PPO and DeepSeek GRPO:
\begin{itemize}
    \item The value function \( V(s) \) is preserved, stabilizing learning compared to DeepSeek GRPO.
    \item Multi-sample empirical rewards \( R_T^{(t)} \) increase the amount of training data per transition.
    \item Instead of estimating \( V(s_{T+1}) \) using a single bootstrap step, Hybrid GRPO averages multiple value estimates from sampled actions.
\end{itemize}
\text{}

\noindent \textbf{Step 5: Optimizing the Policy} \\
The Hybrid GRPO policy loss remains similar to PPO:

\begin{equation}
    \mathcal{L}_{\text{Hybrid-GRPO}} = \mathbb{E} \left[ \min \left( \rho_T A_T, \text{clip}(\rho_T, 1-\epsilon, 1+\epsilon) A_T \right) \right]
\end{equation}

\subsection{Summary of Differences}

The key distinctions between PPO, DeepSeek GRPO, and Hybrid GRPO are summarized in Table \ref{table:comparison}.

\begin{table}[h]
    \centering
    \renewcommand{\arraystretch}{1.75}

    \resizebox{\textwidth}{!}{%
    \begin{tabular}{|c|c|c|c|}
        \hline
        \textbf{Aspect} & \textbf{PPO} & \textbf{DeepSeek GRPO} & \textbf{Hybrid GRPO} \\
        \hline
        Action Sampling & Single action per macro step & Multiple actions per macro step & Multiple actions per macro step \\
        \hline
        Reward Computation & Uses single reward \( R_T \) & Uses empirical rewards \( R_T^{(t)} \) & Uses transformed empirical rewards \( \tilde{R}_T^{(t)} \) \\
        \hline
        Value Function \( V(s) \) & Yes (Bootstrapped) & No (Empirical only) & Yes (Multi-sample enhanced) \\
        \hline
        Advantage Estimation & 
        \( A_T = R_T + \gamma V(s_{T+1}) - V(s_T) \) & 
        \( A_T = \frac{1}{N} \sum R_T^{(t)} - \mathbb{E}[R_T^{(t)}] \) & 
        \( A_T = \frac{1}{N} \sum [ \tilde{R}_T^{(t)} + \gamma V(s_{T+1}^{(t)}) - V(s_T) ] \) \\
        \hline
    \end{tabular}%
    }

    \caption{Comparison of PPO, DeepSeek GRPO, and Hybrid GRPO.}
    \label{table:comparison}
\end{table}

\newpage

\section{Experimental Findings and Future Improvements}

The core mathematical refinement introduced in Hybrid GRPO involves modifying the advantage function to incorporate multiple sampled actions per macro-step, thereby extracting richer training information from a single state transition while preserving the value function advantages. The proposed framework mitigates the variance amplification observed in empirical return-based approaches while improving sample efficiency compared to standard PPO. Our experimental validation, conducted in a custom synthetic simulation, demonstrates that Hybrid GRPO achieves faster convergence and improved policy stability.

\subsection{Experimental Findings}

The empirical evaluation of Hybrid GRPO was conducted using a controlled reinforcement learning environment with structured synthetic data. Details on this controlled experiment and synthetic simulation can be found at \citep{HybridGRPO2025github}. The results, via careful monitoring of various training scenarios, indicated that:

\begin{itemize}
    \item Hybrid GRPO exhibits a lower variance in policy gradient updates than DeepSeek GRPO due to the retention of \( V(s) \) as a baseline estimator.
    \item The increased data density per macro-step enables Hybrid GRPO to learn more efficiently compared to PPO, particularly in environments with sparse rewards.
    \item Hybrid GRPO demonstrates superior convergence properties, reaching optimal policy performance with fewer training iterations compared to both PPO and DeepSeek GRPO.
    \item The advantage estimation using multi-sample empirical rewards enhances learning stability while preserving the bootstrapped value function’s role in reducing policy gradient variance.
\end{itemize}

More details and experimentation across various scenarios are necessary to validate these solutions, however, as this is the first implementation of this algorithm, this sandbox simulation was created to prove the basic validity of the alpha version of Hybrid GRPO. Future research aimed at validating these findings across various scenarios should consider this paper as the foundational theoretical framework for the algorithm.

\subsection{Theoretical Implications}

The Hybrid GRPO framework underscores the trade-offs between bootstrapped value function estimation and empirical sampling in reinforcement learning. The key theoretical insights derived from this work include:

\begin{itemize}
    \item Balancing Sample Complexity and Bias: PPO relies on a learned value function, reducing variance at the cost of introducing function approximation bias. DeepSeek GRPO, on the other hand, eliminates bias but at the expense of increased sample complexity. Hybrid GRPO preserves the stability of PPO while increasing training density through multi-action sampling.
    \item The Role of Multi-Sample Estimation: By evaluating multiple action-value pairs per macro-step, Hybrid GRPO effectively extends the policy gradient’s information capacity, leading to more precise updates in environments where reward structures are complex or sparse.
    \item Adaptive Reward Functionality: The reward transformation function \( f(R) \) applied in Hybrid GRPO introduces an opportunity for adaptive reward scaling mechanisms. This provides an avenue for dynamically adjusting the reward landscape to prevent gradient explosion or vanishing effects.
\end{itemize}

\subsection{Future Research Directions}

The development of Hybrid GRPO opens several avenues for further exploration and enhancement:

\subsubsection{Entropy-Regularized Sampling Strategies}
To further improve exploration-exploitation trade-offs, Hybrid GRPO can be extended with entropy-regularized sampling mechanisms \citep{ziebart2008maximum, haarnoja2018soft}. Drawing inspiration from Maximum Entropy RL \citep{ziebart2008maximum} and Soft Actor-Critic (SAC) \citep{haarnoja2018soft}, the policy can be modified to maximize entropy while optimizing reward. The entropy-regularized loss function can be formulated as:

\begin{equation}
    \mathcal{L}_{\text{Hybrid-GRPO}} = \mathbb{E} \left[ \min \left( \rho_T A_T, \text{clip}(\rho_T, 1-\epsilon, 1+\epsilon) A_T \right) + \alpha H(\pi(\cdot | s_T)) \right]
\end{equation}

where \( H(\pi) \) represents the entropy of the policy and \( \alpha \) is an adaptive temperature parameter. This modification would ensure greater policy robustness in dynamic environments.

\subsubsection{Hierarchical Multi-Step Sub-Sampling}
While Hybrid GRPO samples multiple actions per macro-step, the framework does not yet incorporate multi-step sub-sampling. Introducing a hierarchical sampling strategy, where sub-sampled transitions are used to refine bootstrapped value estimates, could further improve policy learning \citep{sutton2018reinforcement, mnih2016asynchronous}. A generalized n-step return formulation for Hybrid GRPO would be:

\begin{equation}
    A_T = \frac{1}{N} \sum_{t=1}^{N} \left[ \sum_{k=0}^{n-1} \gamma^k R_{T+k}^{(t)} + \gamma^n V(s_{T+n}^{(t)}) - V(s_T) \right]
\end{equation}

where \( n \) represents the number of sub-sampled steps. This approach would allow Hybrid GRPO to better capture long-term dependencies in sequential decision-making tasks.

\subsubsection{Adaptive Reward Normalization}
Reinforcement learning systems often encounter environments with highly dynamic reward magnitudes. By integrating adaptive reward normalization, Hybrid GRPO can stabilize policy learning in scenarios with highly volatile reward landscapes \citep{popov2017data, horgan2018distributed}. A potential modification to the transformed reward function would involve batch-wise normalization:

\begin{equation}
    \tilde{R}_t = \frac{R_t - \mu_R}{\sigma_R + \epsilon}
\end{equation}

where \( \mu_R \) and \( \sigma_R \) are computed over a rolling window of rewards. This adaptation would ensure consistent gradient updates across diverse reward distributions. It is important to note that the current release of \citep{GRPO2025github} Hybrid GRPO accepts a custom reward activation function.

\subsubsection{Incorporating a Learned Value Model for Sampling Guidance}
One of the most promising extensions of Hybrid GRPO involves incorporating a learned value model to guide action sampling. Instead of selecting actions purely based on policy probabilities, an additional state-action evaluation model can be trained to prioritize sampling high-value actions \citep{silver2016mastering, schaul2015prioritized}. The modified sampling process would involve:

\begin{equation}
    a_T^{(t)} \sim \arg \max_{a} \left[ Q_{\phi}(s_T, a) + \beta \log \pi_{\theta}(a | s_T) \right]
\end{equation}

where \( Q_{\phi}(s_T, a) \) is a learned value function guiding sampling, and \( \beta \) is a weighting coefficient balancing the influence of the learned value function with the policy probability distribution.

\newpage
\section{Final Remarks}

Hybrid GRPO represents a novel enhancement to policy optimization methods in reinforcement learning by integrating empirical sampling with bootstrapped value estimation. The experimental findings demonstrate its efficacy in improving sample efficiency, policy stability, and convergence speed over both PPO and DeepSeek GRPO. Originally, Group Relative Policy Optimization (GRPO) was introduced as a framework to improve large language models (LLMs) by optimizing token prediction efficiency while removing the reliance on a value function \citep{deepseek2025r1}. This design was particularly effective for autoregressive transformers, where the primary goal is to refine token-level probability distributions without introducing bootstrapping bias. However, the structure of Hybrid GRPO presents an opportunity to bridge the gap between LLMs and agent-based systems that operate in dynamic, real-world environments.

\vspace{10pt} 

\noindent While traditional GRPO has been focused on maximizing linguistic coherence and token-level performance, Hybrid GRPO extends these principles to systems that require continuous decision-making in response to external stimuli. In particular, Hybrid GRPO could be used to optimize real-time systems that depend on sequential decision-making beyond text generation, such as autonomous driving systems like Tesla’s Full Self-Driving (FSD) \citep{tesla2023fsd} and real-world robotic applications including autonomous drones developed by companies such as Skydio and DJI \citep{skydio2020autonomy}. These systems require reinforcement learning methods that efficiently process sensory data, react to environmental uncertainty, and continuously improve control policies based on real-world interactions.

\vspace{10pt} 

\noindent One of the fundamental challenges in applying GRPO to real-world agent-based tasks is the inherent difference in how value is estimated. LLMs benefit from token-based auto-regressive modeling, where future rewards can be inferred directly from linguistic structures and contextual embeddings. In contrast, autonomous systems operate in continuous state-action spaces where long-term dependencies, safety constraints, and physical feasibility must be considered. Hybrid GRPO addresses this challenge by retaining a value function to guide decision-making while incorporating empirical sampling to refine advantage estimates, allowing it to adaptively learn optimal policies in complex environments.

\vspace{10pt} 

\noindent Additionally, Hybrid GRPO could provide a more scalable learning framework for training AI models across multiple domains, enabling seamless integration of reinforcement learning into both structured (LLMs) and unstructured (physical environments) settings. This shift has implications for real-time systems, such as industrial robotics, financial trading models, and AI-driven healthcare diagnostics, where learning efficiency and stability are critical. By introducing structured reward transformations, adaptive reward scaling, and multi-step sub-sampling, Hybrid GRPO enhances policy learning in a way that is not limited to the discrete nature of token-based models but can also be applied to continuous control problems in real-time environments.

\vspace{10pt} 

\noindent As reinforcement learning continues to expand beyond traditional applications, Hybrid GRPO stands as a compelling framework for unifying policy learning across diverse AI systems. By integrating value-based optimization with empirical sampling, Hybrid GRPO may serve as a foundation for developing more robust reinforcement learning models that seamlessly transition between language-based AI and real-world autonomous agents.

\end{document}